\documentclass[runningheads]{llncs}
\usepackage[T1]{fontenc}
\usepackage{graphicx,verbatim}
\usepackage{amsmath}
\usepackage{amssymb}
\usepackage{xcolor}
\usepackage{booktabs}
\usepackage{multirow} 
\usepackage{threeparttable}
\usepackage{colortbl}
\usepackage{makecell}
\usepackage{subcaption}
\usepackage{hyperref}
\usepackage{marvosym}

\begin{document}
\definecolor{b}{HTML}{00B0F0}
\definecolor{r}{HTML}{DF0000}
\definecolor{c2}{HTML}{FBD9BD}

\title{MrTrack: Register Mamba for Needle Tracking with Rapid Reciprocating Motion during Ultrasound-Guided Aspiration Biopsy}
\titlerunning{MrTrack: Needle Tracking during Aspiration Biopsy}
%

\author{Yuelin Zhang\inst{1} \and
Qingpeng Ding\inst{1} \and 
Long Lei\inst{2} \and
Yongxuan Feng\inst{1} \and \\ 
Raymond Shing-Yan Tang\inst{3} \and 
Shing Shin Cheng\inst{1}\textsuperscript{\Letter}
}

\authorrunning{Y. Zhang et al.}
\institute{$^1$Department of Mechanical and Automation Engineering, \\ 
$^2$Department of Computer Science and Engineering, \\
$^3$Department of Medicine and Therapeutics and Institute of Digestive Disease, \\ 
The Chinese University of Hong Kong} 

\maketitle              
\begin{abstract}

Ultrasound-guided fine needle aspiration (FNA) biopsy is a common minimally invasive diagnostic procedure. However, an aspiration needle tracker addressing rapid reciprocating motion is still missing. MrTrack, an aspiration needle tracker with a mamba-based register mechanism, is proposed. MrTrack leverages a Mamba-based register extractor to sequentially distill global context from each historical search map, storing these temporal cues in a register bank. The Mamba-based register retriever then retrieves temporal prompts from the register bank to provide external cues when current vision features are temporarily unusable due to rapid reciprocating motion and imaging degradation. A self-supervised register diversify loss is proposed to encourage feature diversity and dimension independence within the learned register, mitigating feature collapse. Comprehensive experiments conducted on both robotic and manual aspiration biopsy datasets demonstrate that MrTrack not only outperforms state-of-the-art trackers in accuracy and robustness but also achieves superior inference efficiency. Project page: \href{https://github.com/PieceZhang/MrTrack}{https://github.com/PieceZhang/MrTrack}


\end{abstract}
\section{Introduction}

Ultrasound (US)‐guided fine needle aspiration (FNA) has emerged as a pivotal minimally invasive diagnostic procedure for percutaneous interventions \cite{richardson2017imaging}, facilitating the retrieval of cellular material by rapid reciprocating movement \cite{rumack2017diagnostic}.
Despite its growing clinical importance, a dedicated needle tracker addressing the rapid needle aspiration is still missing.
In practice, the imaging environment during FNA presents unique challenges for a US needle tracker, where reciprocating rapid needle motion during the aspiration \cite{rumack2017diagnostic} induces abrupt nonlinear tip displacement and degraded imaging quality.
Although recent models enhance needle tracking by deep learning \cite{kimbowa2024advancements}, they were mainly designed for stable environments,
yet the nonlinear needle motion and artifact-prone imaging of aspiration biopsy can render these methods ineffective.

Recent developments in tracking reveal that capturing long-range temporal dependencies can enhance tracking \cite{fu2021stmtrack,lai2024mambavt,zheng2024odtrack}. 
However, conventional CNN-based \cite{fu2021stmtrack} or transformer-based \cite{zheng2024odtrack} trackers struggle with computational constraints when integrating temporal context.
In contrast, Mamba \cite{gu2023mamba}, a recently developed framework based on state space models (SSMs) \cite{kalman1960new}, has shown its superiority and efficiency in modeling long-term temporal features with satisfactory complexity growth.
Although there exist Mamba trackers with temporal modeling \cite{lai2024mambavt,xie2024robust,zhang2025mambaxctrack,zhang2024motion}, their performance and efficiency suffer from inefficient temporal integration. 
The register then offers a promising solution \cite{darcet2023vision}, which is a learnable token that aggregates global context into a compact representation for efficient downstream processing. 
It calls for an efficient register mechanism to model a spatiotemporal profile for the aspiration needle. 
The global context from historical search maps should be sequentially \textit{extracted} and stored using the register. The stored latent features from registers should also be effectively \textit{retrieved} to provide external cues for the tracking of the current frame, thus compensating for rapid motion and abrupt appearance changes of the target being tracked.

In this work, \textbf{MrTrack}, an ultrasound needle tracker for FNA based on \textbf{M}amba and \textbf{r}egister, is proposed with a \textbf{register extractor} and a \textbf{register retriever}. 
To the best of our knowledge, it is the first US needle tracker designed for the FNA procedure to robustly track the needle tip undergoing fast reciprocating motion.
A \textbf{register diversify loss (RD loss)} is also proposed as a self-supervised regularization to prevent collapse in the learned register by encouraging register diversity and dimension independence.
The experimental results on both robotic and manual aspiration biopsies show state-of-the-art performance and efficiency of MrTrack.
The contributions of MrTrack are fourfold: 
\textbf{1)} The proposed Mamba-based register extractor sequentially condenses historical frames into compact tokens using a trainable register, storing spatiotemporal profiles to provide external cues from past frames when the needle undergoes appearance variation.
\textbf{2)} The proposed Mamba-based register retriever retrieves temporal cues from historical registers and integrates them with the template map, assisting in tracking an abruptly moving needle using temporal context.
\textbf{3)} The self-supervised register diversify loss (RD Loss) is proposed to prevent feature collapse by promoting diversity and dimension independence, ensuring an informative register that representationally stores the varying temporal context.
\textbf{4)} MrTrack achieves highly efficient inferencing, which enables real-time tracking of rapid needle motion under high-speed US imaging.

\section{Related Work}
\noindent \textbf{Ultrasound needle tracking}
By free-hand \cite{luo2023recon} or robotics-driven \cite{ma2025cross}, US is commonly adopted in many clinical practices \cite{cai2024autonomous,ma2024feasibility}, including US-guided needle insertion.
To track a US needle, 
recent learning-based methods have bee proposed, including discriminative correlation filters \cite{shen2019discriminative}, convolutional neural networks \cite{mwikirize2019single}, and transformer \cite{yan2023learning,yan2024task}. However, these methods were largely developed for relatively stable needle dynamics.
In contrast, the dynamic environment of aspiration biopsy, where rapid needle motion and significant imaging artifacts frequently obscure fine needle features \cite{rumack2017diagnostic}, exposes the limitations of existing trackers. 
Although some needle trackers incorporate historical motion \cite{yan2023learning} to compensate for temporary tip loss, the reciprocating and abrupt movements in aspiration can limit the effectiveness of such motion integration, where the motion can be highly nonlinear and unpredictable.

\noindent \textbf{Mamba} 
Recently, based on SSMs \cite{kalman1960new}, Mamba \cite{gu2023mamba} outperforms transformers, leveraging its selective scan mechanism to integrate information over extended temporal ranges with a linear complexity \cite{zhang2024survey}. 
Mamba-based trackers have demonstrated their capability in tracking tasks and prove the importance of modeling long-term temporal dependencies \cite{lai2024mambavt,xie2024robust}.
In contrast, transformer-based \cite{zheng2024odtrack} or CNN-based \cite{fu2021stmtrack} trackers with temporal modeling tend to struggle under similar computational constraints due to higher complexity.
The existing Mamba trackers with temporal-context modeling face challenges due to the lack of an effective and efficient temporal extraction and integration paradigm \cite{lai2024mambavt,xie2024robust}.
As a trainable prompt, the register has been applied in transformer \cite{darcet2023vision} and Mamba \cite{wang2024mamba} for its ability to suppress artifacts and, more critically, function as a compact global representation.
The learnable register can thus be helpful for aspiration needle tracking by adapting rapid motion with temporal context.

\noindent \textbf{Self-supervised regularization}
Advances in self-supervised learning have demonstrated that enforcing feature invariance can yield powerful representations \cite{gui2024survey,zhang2024unified}. 
Recent approaches like BYOL \cite{grill2020bootstrap}, Barlow Twins \cite{zbontar2021barlow}, and VICReg \cite{bardes2021vicreg} promote feature diversity and prevent collapse through variance preservation and decorrelation of embedding dimensions. 
However, the self-supervised regularization for training learnable prompts like the register in Mamba is still missing.

\section{Method}
\subsection{Preliminaries on Mamba}
The advancement of Mamba is based on the structured state-space models (SSMs), which rely on a continuous linear time-invariant (LTI) system.
It can be expressed with the following linear ordinary differential equation (ODE)
\begin{equation}
\begin{split}
    & h^\prime(t)=\textbf{A}h(t)+\textbf{B}u(t), \\
    & y(t)=\textbf{C}h(t),
    \label{eq_ode}
\end{split}
\end{equation}
where $\textbf{A}\in \mathbb{R}^{N\times N}$, $\textbf{B}\in \mathbb{R}^{N\times 1}$, and $\textbf{C}\in \mathbb{R}^{1\times N}$. 
Following the equations, the input $u(t)\in\mathbb{R}$ is mapped to the output $y(t)\in\mathbb{R}$ by a linear transformation with $h(t)\in\mathbb{R}^{N\times 1}$. 
A discretization is then required for this continuous system.
With $\Delta$ denoting the timescale step size, the discretized system is given by
\begin{align}
\begin{split}
    & h_t=\bar{\textbf{A}}h_{t-1}+\bar{\textbf{B}}u_t, \\
    & y_t=\textbf{C}h_t,
    \label{eq_ode_disc}
\end{split}
\end{align}
where $\textbf{A}$ and $\textbf{B}$ is discretized to be $\bar{\textbf{A}}=exp({\Delta}\cdot \textbf{A})$ and $\bar{\textbf{B}}=({\Delta}\cdot \textbf{A})^{-1}(exp({\Delta}\cdot\textbf{A})-I)\cdot {\Delta}\cdot\textbf{B}$.
In Mamba \cite{gu2023mamba}, the selective scan mechanism is incorporated to learn $\textbf{B},\textbf{C},\Delta$ from input $x$ by an input-dependent way, which are given by $\textbf{B}=\text{Linear}(x)$, $\textbf{C}=\text{Linear}(x)$, $\Delta=\text{Softplus}(\widetilde{\Delta}+\text{Linear}(x))$, where $\widetilde{\Delta}$ is learnable.

\begin{figure*}
\centering
\includegraphics[width=\textwidth]{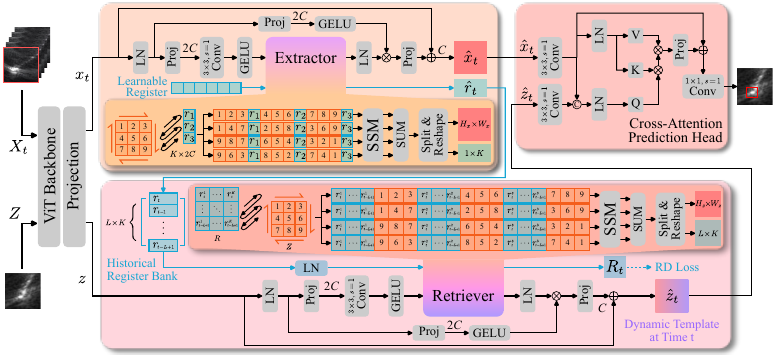}
\caption{
Overview of MrTrack. The notes (e.g. $3\times3, s=1$) on convolution layers represent the kernel size and stride.
}
\label{fig_overview}
\end{figure*}

\subsection{Overview of MrTrack}
As shown in Fig.~\ref{fig_overview}, a ViT-Base \cite{dosovitskiy2020image} is adopted as the backbone, where the embedded features of the template map $z\in \mathbb{R}^{H_z \times W_z \times C}$ and search map $x_t\in \mathbb{R}^{H_x \times W_x \times C}$ ($C=256$) are obtained, where $t$ denotes the current time step.
The $x_t$ and $z$ are then passed to the Mamba-based register extractor and retriever, respectively, to extract registers from the search map and retrieve information into the template map.
The register is trained in a self-supervised manner by the register diversify loss $\mathcal{L}_{\mathrm{RD}}$.
With the obtained $\hat{x}_t$ and $\hat{z}_t$, the tracking result is predicted by the cross-attention prediction head.

\subsection{Mamba-based Register Extractor and Register Retriever}
The proposed Mamba-based register extraction-retrieval framework benefits aspiration needle tracking by aggregating long-term visual features at minimal overhead.
The extractor learns compact global representation from a search map $x_t$ through a trainable register $r\in \mathbb{R}^{k \times 2C}$ ($k=8$), where $r$ is shareable for all frames. 
Specifically, after a series of layer normalization, projection, and convolution, $x_t$ is passed to the extractor together with $r$, in which a cross-map scanning is performed to evenly insert the register \textit{behind} each image segment throughout the $x_t$, as shown in Fig.~\ref{fig_overview}.
This insertion ensures a thorough feature translation.
The fused vectors are then learned by the SSM, where not only the feature from $x_t$ is conveyed into the output $r_t$ but also the $x_t$ itself is modeled to be $\hat{x}_t$ with richer representation and suppressed artifacts \cite{wang2024mamba}.
By the extractor, the dedicated compact descriptor for each frame is efficiently obtained, then stored in the register bank $R_t = [r_t, ..., r_{t-L+1}] \in \mathbb{R}^{L \times k \times 2C}$, $L=300$.

The retriever retrieves the information from the register bank $R_t$ to the template $z$.
Through cross-map scanning, the descriptors $r_i$ in $R_t$ are evenly inserted back into the sequence of image tokens of $z$, yet \textit{before} each image segment. 
After SSM modeling, the temporal context in $R_t$ is retrieved into $\hat{z}_t$, which is a dynamic template containing global temporal context from time $t$ to $t-L$.
Through the retriever, the feature of the initial template is aggregated with the latest transient cues, which can be helpful during aspiration when abrupt and rapid motion happens.

\subsection{Register Diversify Loss}
To ensure that the learned register does not collapse and to encourage distinctiveness in the register bank, the register diversify loss (RD loss) is proposed. 
RD loss comprises two components: a variance promoting term $\mathcal{L}_{var}$ and a cross-register diversify term $\mathcal{L}_{div}$. The variance promoting term $\mathcal{L}_{var}$ penalizes feature dimensions whose standard deviation falls below a threshold $\tau$ ($\tau=1$ in this paper) to encourage diversity. 
The \textit{softplus} is applied here as a smooth penalty with differentiability, given by $\text{softplus}(v)=ln(1+e^v)$.
$\mathcal{L}_{var}$ is then defined by
\begin{align}
\begin{split}
    L_{var} = \frac{1}{d}\sum_{i=1}^{d}\text{softplus}(\tau - \sqrt{V_i + \epsilon}),
    \label{eq_L_var}
\end{split}
\end{align}
where $d$ is the dimensionality, $\epsilon$ is a small constant, $V_i$ is variance of the $i_{\text{th}}$ dimension. 
The variance of the learnable register is then promoted with $\mathcal{L}_{var}$.

Except for promoting variance of the register, the cross-register diversify term $\mathcal{L}_{div}$ is proposed to encourage the diversity within the register bank, thus preventing the register extractor from encoding similar features for different frames.
Assume each descriptor $r_i$ in $R_t = [r_t, ..., r_{t-L}]$ is defined by $r_i=[r_{i,1},...,r_{i,d}]\in \mathbb{R}^{k \times d}$, where $d=2C$, $\mathcal{L}_{div}$ is presented with a covariance format given by
\begin{align}
\begin{split}
    \mathcal{L}_{\mathrm{div}} = \frac{1}{dL(L-1)} \sum_{p=1}^{d} \sum_{\substack{i,j=1 \\ i \neq j}}^{L} \left[ \left(r_{i,p} - \bar{r}\right) \left(r_{j,p} - \bar{r} \right) \right]^2,~\text{where}~\bar{r}=\frac{1}{L}\sum_{n=1}^{L} r_{n,p}.
    \label{eq_L_div}
\end{split}
\end{align}
$\mathcal{L}_{div}$ drives the off-diagonal elements of the covariance close to zero thus promoting decorrelation.
Different from the covariance regularization in \cite{bardes2021vicreg} that decorrelates the dimension within the same sample, $\mathcal{L}_{div}$ decorrelates dimensions between different samples $r_i$ from the register bank, thus encouraging the model to learn independently varying features for different frames. By combining Eq.~\ref{eq_L_var} and Eq.~\ref{eq_L_div} with scalar $\alpha=\beta=0.01$, the RD loss $\mathcal{L}_{\mathrm{RD}}$ is given by
\begin{align}
\begin{split}
    \mathcal{L}_{\mathrm{RD}} = \alpha \mathcal{L}_{\mathrm{var}} + \beta \mathcal{L}_{\mathrm{div}}.
    \label{eq_rdl}
\end{split}
\end{align}
This self-supervised regularization prevents collapse by encouraging dimension independence in the learnable register and promoting diversity in register bank.

\begin{figure*}
\centering
\includegraphics[width=\textwidth]{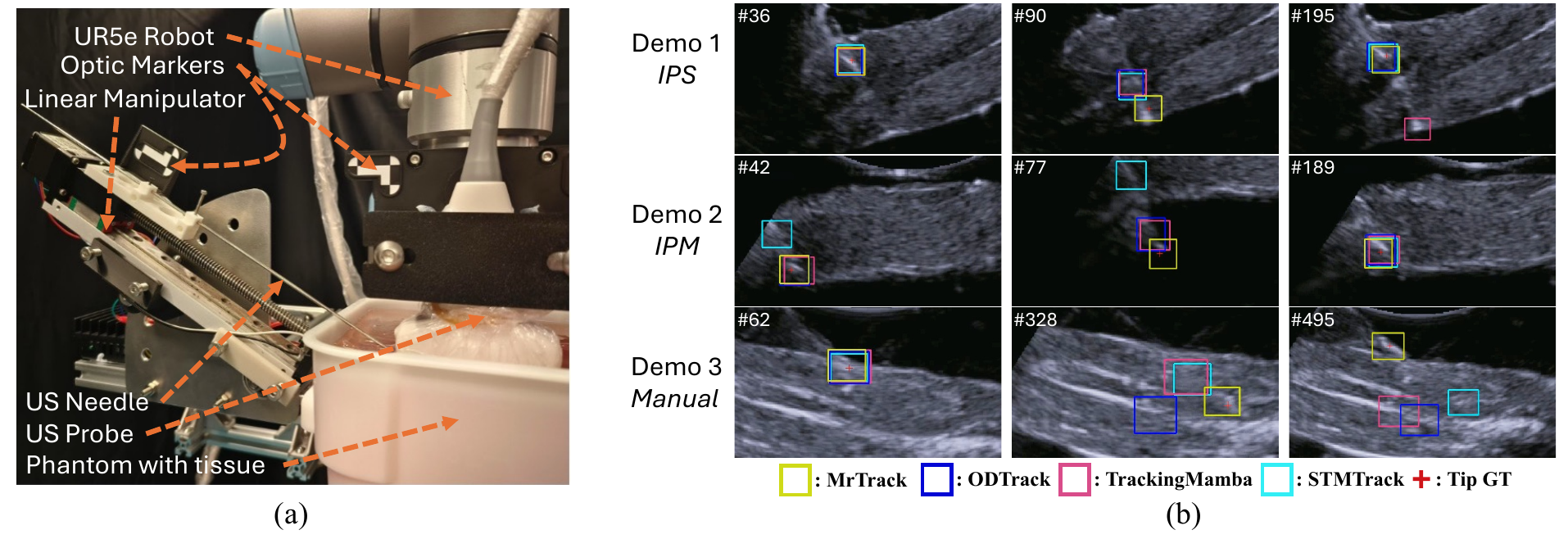}
\caption{
Experiment setup (a) and tracking demonstration (b). More demonstrations are shown in the supplementary video.
}
\label{fig_exp}
\end{figure*}

\section{Experiments and Results}
\subsection{Experiment Setup and Data Collection}
As shown in Fig.~\ref{fig_exp}(a), a robotic US needle insertion platform and a US imaging system with a Wisonic Clover 60 US machine and a Wisonic C5-1 convex transducer (sampling frequency 4.5 MHz) was adopted for the experiments.
A 20 gauge fine aspiration needle \cite{d2016ultrasound} was used to insert into a phantom made with fresh pork supported by solidified agar. 
The ground truth position of the needle tip was collected by an optic tracker (ClaroNav MicronTracker 3), which has an RMSE of 0.189 mm in our setup.
Datasets for both robotic and manual needle manipulation experiments simulating the aspiration biopsy procedure were collected.
During the robotic insertion, the needle was installed on a linear manipulator, while the US probe was held by a UR5e robot arm.
The needle was first inserted at a constant velocity of 20 mm/s to reach the target with a depth of more than 50 mm. After reaching the target, the needle would move back and forth five times at a velocity of 30 mm/s and a depth of 15 mm to perform aspiration.
Both in-plane-static (IPS) and in-plane-moving (IPM) techniques \cite{kimbowa2024advancements,yan2023learning} were performed at three needle insertion angles ($30^\circ$, $45^\circ$, $60^\circ$).
The robotic insertion dataset includes 54,428 frames in videos ($1920\times1080$, 30 FPS) from 239 trials (105 IPS, 134 IPM). After sampling it by 5 to remove redundancy, the dataset was divided into training (7,619 samples), validation (1,088 samples), and testing (2,177 samples) sets in the ratio 7:1:2.
During the manual insertion, both the needle and probe were held by the user. Four users were asked to perform four aspirations each, with the needle being moved back and forth two to three times per second. 
It resulted in a dataset that includes 16 videos (10,948 frames), which is only for testing to validate generalization.

During the evaluation, all models were implemented using PyTorch and trained on the same dataset with four NVIDIA RTX 4090 GPUs.
All models adopt the same training strategy (350 epochs, batch size 32, AdamW optimizer) and the same input size of $384\times384$ and $192\times192$ for search and template map. Scaling, blur, and position shifting were adopted for image augmentation. The learning rate was set to 5e-4 and dropped by 10 after 200 epochs.

\begin{table*}
\caption{
Evaluation and ablation study results. The methods with the best and the second best performance are in {\color{r}red} and {\color{b}cyan}. 
AUC and \textit{P} are reported in percentage (\%), and Err and SD are reported in mm.
In ablation studies, the performance difference relative to the baseline has been highlighted.
}
\label{tab_exp}

\begin{subtable}[b]{0.8\textwidth}
\centering
\setlength{\tabcolsep}{0.35mm}{
\scalebox{0.75}{
\begin{tabular}{lccccccccc|ccc|c}
\toprule
\multirow{3}{*}{Method} & \multicolumn{9}{c|}{\textbf{Robotic}}  & \multicolumn{3}{c|}{\multirow{2}{*}{\textbf{Manual}}} & \multirow{3}{*}{FPS}\\
& \multicolumn{3}{c}{In-plane-static (IPS)}  & \multicolumn{3}{c}{In-plane-moving (IPM)} & \multicolumn{3}{c|}{Mean} & \multicolumn{3}{c|}{} & \\
\cmidrule(lr){2-13}
& \cellcolor{c2!50}AUC$_\uparrow$ & \cellcolor{c2!50}$\textit{P}$$_\uparrow$ & \cellcolor{c2!50}Err$\pm$SD$_\downarrow$ & \cellcolor{c2!50}AUC$_\uparrow$ & \cellcolor{c2!50}$\textit{P}$$_\uparrow$ & \cellcolor{c2!50}Err$\pm$SD$_\downarrow$ & \cellcolor{c2!50}AUC$_\uparrow$ &  \cellcolor{c2!50}$\textit{P}$$_\uparrow$ & \cellcolor{c2!50}Err$\pm$SD$_\downarrow$ & \cellcolor{c2!50}AUC$_\uparrow$ & \cellcolor{c2!50}$\textit{P}$$_\uparrow$ & \cellcolor{c2!50}Err$\pm$SD$_\downarrow$ \\
\midrule
SiamRPN++ \cite{li2019siamrpn++} & 44.1 & 60.0 & 4.71$\pm$3.10 & 58.2 & 75.0 & 3.28$\pm$3.00 & 50.3 & 66.7 & 4.14$\pm$3.06 & 42.0 & 59.3 & 5.50$\pm$3.73 & 31.3 \\
SiamCAR \cite{guo2020siamcar} & 45.3 & 68.8 & 4.20$\pm$2.38 & 59.7 & 80.5 & 2.84$\pm$2.25 & 52.0 & 74.6 & 3.66$\pm$2.33 & 40.3 & 57.1 & 5.72$\pm$3.49 & 48.2 \\
SiamBAN \cite{chen2022siamban} & 52.9 & 73.1 & 3.90$\pm$2.51 & 61.5 & 89.8 & 2.39$\pm$2.09 & 56.2 & 79.6 & 3.30$\pm$2.34 & 49.0 & 69.5 & 3.60$\pm$3.18 & 33.2 \\
SwinTrack \cite{lin2022swintrack} & 51.0 & 72.6 & 3.93$\pm$1.83 & \color{b}64.4 & 93.1 & 1.91$\pm$1.72 & 57.1 & 80.6 & 3.13$\pm$1.79 & 48.2 & 70.6 & 3.62$\pm$2.85 & 60.1 \\
STMTrack \cite{fu2021stmtrack} & 54.7 & 76.9 & 3.80$\pm$2.00 & 64.1 & 92.5 & 2.10$\pm$1.98 & 59.3 & 84.6 & 3.14$\pm$2.00 & 51.5 & 76.8 & {\color{b}3.05}$\pm$2.42 & 24.4 \\
TrackingMamba \cite{wang2024trackingmamba} & 55.2 & \color{b}79.7 & \color{b}3.62$\pm$1.71 & 63.9 & 92.9 & 1.90{\color{b}$\pm$1.62} & 59.5 & \color{b}86.3 &{\color{b}2.93}$\pm$1.72 & 54.1 & \color{b}77.0 & 3.39$\pm$2.98 & \color{b}63.1 \\
ODTrack \cite{zheng2024odtrack} & \color{b}57.6 & 78.6 & 3.76$\pm$1.73 & 63.5 & \color{r}94.5 & {\color{r}1.81}$\pm$1.69 & \color{b}60.8 & 85.0 & 2.98{\color{b}$\pm$1.71} & \color{b}55.4 & 76.2 & 3.20{\color{b}$\pm$2.11} & 53.2 \\
MrTrack & \color{r}63.6 & \color{r}92.0 & \color{r}2.51$\pm$1.31 & \color{r}67.2 & \color{b}93.2 & {\color{b}1.88}{\color{r}$\pm$1.10} & \color{r}65.4 & \color{r}92.3 & \color{r}2.29$\pm$1.22 & \color{r}59.0 & \color{r}83.9 & \color{r}2.80$\pm$1.59 & \color{r}73.9 \\
\midrule
\end{tabular}
}
}
\end{subtable}

\begin{subtable}[b]{0.8\textwidth}
\centering
\setlength{\tabcolsep}{0.42mm}{
\scalebox{0.722}{
\begin{tabular}{lccc|ccc|c}
\midrule
\multirow{3}{*}{Ablations} & \multicolumn{3}{c|}{\textbf{Robotic (Mean)}}  & \multicolumn{3}{c|}{\textbf{Manual (Mean)}} & \multirow{2}{*}{FPS} \\
\cmidrule(lr){2-7}
& \cellcolor{c2!50}AUC$_\uparrow$ & \cellcolor{c2!50}$\textit{P}$$_\uparrow$ & \cellcolor{c2!50}Err$\pm$SD$_\downarrow$ & \cellcolor{c2!50}AUC$_\uparrow$ & \cellcolor{c2!50}$\textit{P}$$_\uparrow$ & \cellcolor{c2!50}Err$\pm$SD$_\downarrow$ \\
\midrule
Baseline: MrTrack & 65.4 & 92.3 & 2.29$\pm$1.22 & 59.0 & 83.9 & 2.80$\pm$1.59 & 73.9\\
$v_{r1}$: $L=100,k=8$ & 63.5{\scriptsize(-1.9)} & 92.2{\scriptsize(-0.1)} & 2.27{\scriptsize(-0.02)}$\pm$1.73{\scriptsize(+0.51)} & 59.7{\scriptsize(+0.7)} & 82.1{\scriptsize(-1.8)} & 2.94{\scriptsize(+0.14)}$\pm$1.72{\scriptsize(+0.13)} & 76.2{\scriptsize(+2.3)} \\
$v_{r2}$: $L=\infty,k=8$ & 65.9{\scriptsize(+0.5)} & 94.0{\scriptsize(+1.7)} & 2.19{\scriptsize(-0.10)}$\pm$1.34{\scriptsize(+0.12)} & 57.8{\scriptsize(-1.2)} & 80.6{\scriptsize(-3.3)} & 2.94{\scriptsize(+0.14)}$\pm$1.80{\scriptsize(+0.21)} & 58.0{\scriptsize(-15.9)} \\
$v_{r3}$: $L=300,k=2$ & 65.1{\scriptsize(-0.3)} & 92.2{\scriptsize(-0.1)} & 2.24{\scriptsize(-0.05)}$\pm$1.25{\scriptsize(+0.03)} & 58.1{\scriptsize(-0.9)} & 82.9{\scriptsize(-1.0)} & 2.92{\scriptsize(+0.12)}$\pm$1.55{\scriptsize(-0.04)} & 73.7{\scriptsize(-0.2)} \\
$v_{r4}$: $L=300,k=32$ & 62.4{\scriptsize(-3.0)} & 91.2{\scriptsize(-1.1)} & 2.31{\scriptsize(+0.02)}$\pm$1.39{\scriptsize(+0.17)} & 53.8{\scriptsize(-5.2)} & 81.9{\scriptsize(-2.0)} & 3.19{\scriptsize(+0.39)}$\pm$1.99{\scriptsize(+0.40)} & 68.0{\scriptsize(-5.9)} \\
$v_{M1}$: \textit{Trans} \textit{E} \& \textit{R} & 59.0{\scriptsize(-6.4)} & 85.8{\scriptsize(-6.5)} & 3.10{\scriptsize(+0.81)}$\pm$1.40{\scriptsize(+0.18)} & 55.3{\scriptsize(-3.7)} & 78.8{\scriptsize(-5.1)} & 2.99{\scriptsize(+0.19)}$\pm$1.65{\scriptsize(+0.06)} & 61.3{\scriptsize(-12.6)} \\
$v_{M2}$: w/o Register & 61.2{\scriptsize(-4.2)} & 88.5{\scriptsize(-3.8)} & 2.49{\scriptsize(+0.20)}$\pm$1.79{\scriptsize(+0.57)} & 54.2{\scriptsize(-4.8)} & 79.4{\scriptsize(-4.5)} & 3.38{\scriptsize(+0.58)}$\pm$2.03{\scriptsize(+0.44)} & 76.1{\scriptsize(+2.2)} \\
$v_{RDL}$: w/o $\mathcal{L}_{\mathrm{RD}}$ & 64.1{\scriptsize(-1.3)} & 90.5{\scriptsize(-1.8)} & 2.40{\scriptsize(+0.11)}$\pm$1.65{\scriptsize(+0.43)} & 56.4{\scriptsize(-2.6)} & 83.5{\scriptsize(-0.4)} & 2.88{\scriptsize(+0.08)}$\pm$1.77{\scriptsize(+0.18)} & 73.9{\scriptsize(-0.0)} \\
\bottomrule
\end{tabular}
}
}
\end{subtable}

\end{table*}

\subsection{Results and Discussion}
The evaluation was performed between MrTrack and seven trackers, including classic Siamese trackers \cite{chen2022siamban,guo2020siamcar,li2019siamrpn++}, a tracker with motion integration \cite{lin2022swintrack}, a Mamba-based tracker \cite{wang2024trackingmamba}, and trackers with temporal modeling \cite{fu2021stmtrack,zheng2024odtrack}.
The results are reported by four metrics, including average error (Err) and standard deviation (SD) in mm, and commonly used tracking evaluation metrics area under curve (AUC) \cite{wu2013online} and precision ($\textit{P}$) \cite{muller2018trackingnet} in percentage (\%). The inference speed (FPS) is also reported.
As shown in Tab.~\ref{tab_exp}, MrTrack achieves the best performance for most metrics while maintaining the best efficiency with an inference speed of 73.9 FPS.
In Err for robotic and manual insertion, it outperforms the second-best method by 21.8\% and 8.2\%, respectively. For SD, the improvements are even more substantial, reaching 28.7\% and 24.6\%. 
For manual insertion, which involves more abrupt motion and imaging disturbances, MrTrack still achieves consistent SOTA performance, which shows its robustness and generalizability to unseen images and motion, as the manual insertion data was not used for training. 
The demonstration in Fig.~\ref{fig_exp}(b) also shows that MrTrack achieves the most accurate tracking of the aspiration needle with reciprocating motion.
More examples are provided in the supplementary video.

This result demonstrates the effectiveness of MrTrack with the Mamba-based register, which captures the global spatiotemporal profile of a rapidly moving needle during aspiration.
Among the comparison methods, ODTrack \cite{zheng2024odtrack} also adopts temporal learning yet with worse performance and lower FPS. It may be due to its \textit{token propagation} strategy, which with a fixed-length token to record the temporal context and pass through the whole network, leading to error accumulation.
In contrast, MrTrack extracts and stores each frame individually, ensuring that every historical token possesses the same importance, thereby preventing error accumulation.
Although TrackingMamba \cite{wang2024trackingmamba} also adopts a Mamba framework, it performs poorly because it only utilizes the vanilla Mamba without temporal context modeling.
STMTrack \cite{fu2021stmtrack} also learns temporal features by simply concatenating historical frames but has lower FPS and worse performance due to the lack of an effective temporal integration strategy.
SwinTrack \cite{lin2022swintrack} adopts a historical motion fusion strategy to model temporal context, but since the motion in aspiration is nonlinear and abrupt, this strategy may lead to worse performance for this task.
Currently, MrTrack still lacks a mechanism to update the template when needle visibility deteriorates. Future work will introduce an explicit template updating strategy to address this issue.

\subsection{Ablation Study}
Ablation studies were carried out on 7 variations. 
$v_{r1}$ to $v_{r4}$ explore the temporal integration capacity of the register, where the results indicate that the baseline structure ($L=300,k=8$) with a moderate register bank length $L$ and a reasonable register scale $k$ is the most suitable for the aspiration needle tracking task. 
Although $v_{r2}$ somewhat improves performance by integrating infinite registers, it also leads to a significant decrease in inference speed, as well as a loss of robustness caused by error accumulation.
$v_{M1}$ replaces the proposed Mamba-based register extractor and retriever with a fundamental self-attention transformer \cite{dosovitskiy2020image} similar to \cite{darcet2023vision}.
This results in degradation across all metrics, indicating that the Mamba-based register framework is more effective than transformer.
$v_{M2}$ further shows the effectiveness of the register-based temporal integration, where the register and register bank are removed while keeping Mamba only.
$v_{RDL}$ only applies supervised training without $\mathcal{L}_{\mathrm{RD}}$, where the result shows the effectiveness, i.e., register diversity and dimension independence are promoted, thus enhancing the whole register framework and aspiration needle tracking performance.
The robustness of $\mathcal{L}_{\mathrm{RD}}$ is also showed by avoiding tuning the value of hyper-parameters $\alpha$ and $\beta$.

The results demonstrate the effectiveness of the proposed modules under rapid motion and degraded image quality, where both the learnable register mechanism and the register diversify loss are crucial for robust temporal integration and feature discrimination. 
It underscores the importance of capturing long-term temporal context, which not only improves tracking accuracy but also enhances stability across various insertion techniques.

\section{Conclusion}
In this paper, a US needle tracker, MrTrack, has been proposed for FNA which features fast reciprocating needle motion. It leverages a Mamba-based register mechanism and a self-supervised register diversify loss to ensure robust and accurate needle tip tracking. 
In the future, more experiments will be conducted on large-scale manual insertion data and real clinical video, as well as on handling needle out-of-plane caused by deflection.

    

\begin{credits}
\subsubsection{\ackname} 
Research reported in this work was supported in part by Research Grants Council (RGC) of Hong Kong (T45-401/22-N, CUHK 14217822, CUHK 14207823, and AoE/E-407/24-N) and in part by Innovation and Technology Commission of Hong Kong (ITS/234/21, ITS/235/22,  ITS/224/23,  ITS/225/23, and Multi-scale Medical Robotics Center, InnoHK). The content is solely the responsibility of the authors and does not necessarily represent the official views of the sponsors.

\subsubsection{\discintname}
The authors have no competing interests to declare that are relevant to the content of this article
\end{credits}

%
%

\bibliographystyle{splncs04}
\bibliography{reference}

\end{document}